\documentclass{article}

     \PassOptionsToPackage{numbers, compress}{natbib}
     \usepackage[final]{neurips_2020}

\usepackage[utf8]{inputenc} % allow utf-8 input
\usepackage[T1]{fontenc}    % use 8-bit T1 fonts
\usepackage{hyperref}       % hyperlinks
\usepackage{url}            % simple URL typesetting
\usepackage{booktabs}       % professional-quality tables
\usepackage{amsfonts}       % blackboard math symbols
\usepackage{nicefrac}       % compact symbols for 1/2, etc.
\usepackage{microtype}      % microtypography
\usepackage{amsmath}
\usepackage{amssymb,bm,graphicx}
\usepackage[ruled,vlined]{algorithm2e}

\DeclareMathOperator{\tensorize}{ten}
\DeclareMathAlphabet\mathcalbf{OMS}{cmsy}{b}{n}

\title{Multi-Graph Tensor Networks }

\author{%
  Yao Lei Xu, Kriton Konstantinidis, Danilo P. Mandic \\
  Department of Electrical and Electronic Engineering\\
  Imperial College London\\
  \texttt{\{yao.xu15,k.konstantinidis19,d.mandic\}@imperial.ac.uk}}

\begin{document}

\maketitle

\begin{abstract}

The irregular and multi-modal nature of numerous modern data sources poses serious challenges for traditional deep learning algorithms. To this end, recent efforts have generalized existing algorithms to irregular domains through graphs, with the aim to gain additional insights from data through the underlying graph topology. At the same time, tensor-based methods have demonstrated promising results in bypassing the bottlenecks imposed by the Curse of Dimensionality. In this paper, we introduce a novel \textit{Multi-Graph Tensor Network} (MGTN) framework, which exploits both the ability of graphs to handle irregular data sources and the compression properties of tensor networks in a deep learning setting. The potential of the proposed framework is demonstrated through an MGTN based deep Q agent for Foreign Exchange (FOREX) algorithmic trading. By virtue of the MGTN, a FOREX currency graph is leveraged to impose an economically meaningful structure on this demanding task, resulting in a highly superior performance against three competing models and at a drastically lower complexity.
  
\end{abstract}

\section{Introduction}

Deep learning techniques have been at the core of machine learning research for over a decade and have proved successful in a number of areas, including computer vision and speech processing \cite{zhang2018survey}. However, as we enter the era of Big Data, the associated multi-modal and irregular nature of data is posing stern challenges to traditional learning systems, also owing to the sheer volume, variety, veracity and velocity of modern data sources \cite{Cichocki2014}. To this end, it is necessary to generalize deep learning approaches to handle such irregular and multi-modal data.
% Developments in neural networks suggest that, despite the overwhelming success, classical deep learning models such as convolutional neural networks do not perform well for Big Data applications that reside on irregular domains \cite{bronstein2017geometric}. 

Some of the most successful approaches for data analytics on irregular domains resort to graph signal processing techniques, because of their ability to provide insights into both the data acquisition process and their generative mechanisms \cite{shuman2013emerging}. By virtue of their ability to account for the underlying data structure, graph-based learning algorithms have proved advantageous in applications where the graph is known a-priori \cite{wu2020comprehensive}. When it comes to exceedingly large multi-modal data, tensor-based methods have demonstrated their potential in areas including multi-modal learning \cite{7038247}, compression of large-dimensional data \cite{cichocki2016tensor}, and interpretability of neural networks \cite{cohen2016expressive}. In particular, tensor decomposition (TD) and tensor networks (TN) leverage the multi-modality of many Big Data applications to compress large-dimensional data while preserving their structure and interpretability, thus effectively bypassing the bottlenecks imposed by the Curse of Dimensionality \cite{Cichocki2014}.

Despite progress in these individual fields, the full potential arising from the combination of tensors, graphs, and neural networks has only begun to be explored. One such recent approach is the \textit{Recurrent Graph Tensor Network} (RGTN) \cite{xu2020recurrent}, which provides a framework for modelling multi-modal sequential data, through a unifying account of the expressive power of graphs and tensor networks. The initial RGTN model has been introduced primarily for sequential data and is therefore only defined on a single graph domain, often impractical for Big Data applications. 

To provide a general framework that fully exploits the advantages of both graphs and tensors in a deep learning setting, we here generalize the RGTN concept in \cite{xu2020recurrent} to introduce the novel \textit{Multi-Graph Tensor Network} (MGTN). In this way, the proposed MGTN is capable of handling irregular data residing on multiple graph domains, while simultaneously leveraging the compression properties of tensor networks to enhance modelling power and reduce parameter complexity.

The proposed model is verified on the task of Foreign Exchange (FOREX) algorithmic trading, a notoriously challenging paradigm characterized by highly irregular and noisy data \cite{de2020machine}. More specifically, we consider a deep Q trading agent \cite{mnih2013playing}, where an MGTN is used to approximate the action values. By combining the advantages of graphs, tensors, and neural networks, the proposed MGTN agent is shown to yield highly superior performance against three competing agents. This setting is general enough to suggest the use of MGTN in a range of other application domains, including social networks, communication networks, and cognitive neuroscience.

The rest of the paper is organized as follows. We first briefly present the theoretical background necessary to follow this work in Section \ref{sec:Prelim}. The MGTN framework is introduced in Section \ref{sec:GTN}, followed by an in-depth analysis of how FOREX algorithmic trading can be naturally modelled with an MGTN in Section \ref{sec:FE}. Numerical experiments are provided in Section \ref{sec:exp} and Conclusion along with promising future research directions and potential MGTN application domains are given in Section \ref{sec:conc}.

\section{Preliminaries}
\label{sec:Prelim}

\subsection{Tensors and Tensor Networks}

A real-valued tensor is a multidimensional array, denoted by a calligraphic font, e.g., $\mathcalbf{X} \in\mathbb{R}^{I_1\times\dots\times I_N}$, where $N$ is the order of the tensor and $I_n$ ($1 \leq n \leq N$) the size of its $n$\textsuperscript{th} mode. Matrices (denoted by bold capital letters, e.g., $\mathbf{X}\in\mathbb{R}^{I_1\times I_2}$) can be seen as order-2 tensors ($N=2$), vectors (denoted by bold lower-case letters, e.g., $\mathbf{x}\in\mathbb{R}^{I}$) can be seen as order-1 tensors ($N=1$), and scalars (denoted by lower-case letters, e.g., $x\in\mathbb{R}$) are tensors of order $N=0$. A specific entry of a tensor $\mathcalbf{X}\in\mathbb{R}^{I_1\times\dots\times I_N}$ is given by $x_{i_1,\dots,i_N}\in\mathbb{R}$. The tensor indices in this paper are grouped according to the Little-Endian convention \cite{Dolgov2014}.

\textbf{Kronecker Product} A (left) Kronecker product between two tensors, $\mathcalbf{A} \in \mathbb{R}^{I_1 \times \cdots \times I_N}$ and $\mathcalbf{B} \in \mathbb{R}^{J_1 \times \cdots \times J_N}$, denoted by $\otimes$, yields a tensor  $\mathcalbf{C} \in \mathbb{R}^{I_1 J_1 \times \cdots \times I_N J_N}$, of the same order, with entries $c_{\overline{i_1j_1},\ldots,\overline{i_Nj_N}} = a_{i_1, \ldots, i_N} b_{j_1, \ldots, j_N}$, where $\overline{i_n j_n} = j_n + (i_n - 1) J_n$ \cite{Cichocki2014}.

\textbf{Matricization and Tensorization} The mode-$n$ matricization of a tensor $\mathcalbf{X}\in\mathbb{R}^{I_1\times\cdots\times I_N}$ reshapes the multidimensional array into a matrix $\mathbf{X}_{(n)}\in\mathbb{R}^{I_n\times I_1I_2\cdots I_{n-1}I_{n+1}\cdots I_N}$ with $(x_{(n)})_{i_n,\overline{i_1\dots i_{n-1}i_{n+1}\dots i_N}}=x_{i_1,\dots,i_N}$. The inverse process, \textit{Tensorization}, is denoted by $\tensorize(\cdot)$.

% \paragraph{Tensor Contraction}
\textbf{Tensor Contraction} An $(m,n)$-contraction \cite{Cichocki2014} denoted by $\times^m_n$, between an order-$N$ tensor $\mathcalbf{A} \in \mathbb{R}^{I_1\times \cdots \times I_n \times \cdots \times I_N}$ and an order-$M$ tensor $\mathcalbf{B}\in \mathbb{R}^{J_1\times \dots \times J_m \times \dots \times J_M} $, where $I_n = J_m$, yields a third order-$(N+M-2)$ tensor, $\mathcalbf{C}\in \mathbb{R}^{I_1 \times \cdots \times I_{n-1} \times I_{n+1}  \times \cdots \times I_N \times J_1 \times \cdots \times J_{m-1} \times J_{m+1}  \times \cdots \times J_M}$, where $c_{i_1,\dots,i_{n-1}, i_{n+1}, \dots, i_N, j_1, \dots, j_{m-1}, j_{m+1}, \dots, j_M} = \sum_{i_n=1}^{I_n} a_{i_1, \dots, i_{n-1}, i_n, i_{n+1}, \dots, i_N } b_{j_1, \dots, j_{m-1}, i_n, j_{m+1}, \dots, j_M}$. 
% \begin{equation}\label{eq:cont}
% 	\begin{aligned}
% 		&c_{i_1,\dots,i_{n-1}, i_{n+1}, \dots, i_N, j_1, \dots, j_{m-1}, j_{m+1}, \dots, j_M   } \\
% 		&= \sum_{i_n=1}^{I_n} a_{i_1, \dots, i_{n-1}, i_n, i_{n+1}, \dots, i_N } b_{j_1, \dots, j_{m-1}, i_n, j_{m+1}, \dots, j_M}   
% 	\end{aligned}
% \end{equation}

\textbf{Tensor Network}
A Tensor Network (TN) is a tensor architecture comprised of smaller-order core tensors which are connected by tensor contractions, where each tensor is represented as a node, while the number of edges that extends from that node corresponds to tensor order \cite{cichocki2016tensor}. If two nodes are connected through an edge, it represents a linear contraction between two tensors over modes of equal dimensions. 

\textbf{Tensor Decomposition} 
Special instances of tensor networks include those based on Tensor Decomposition (TD) methods, which approximate high-order, large-dimension tensors via contractions of smaller core tensors, therefore drastically reducing the computational complexity in tensor manipulation while preserving the data structure \cite{cichocki2016tensor}. We here consider the Tensor-Train decomposition (TTD) \cite{oseledets2011tensor}, a highly efficient TD method that can decompose a large order-$N$ tensor, $\mathcalbf{X} \in \mathbb{R}^{I_1 \times I_2 \times \cdots \times I_N}$, into interconnected smaller core tensors, $\mathcalbf{G}^{(n)} \in \mathbb{R}^{ R_{n-1} \times  I_n \times R_n }$, as $\mathcalbf{X} = \mathcalbf{G}^{(1)} \times^1_2 \mathcalbf{G}^{(2)} \times^1_3 \mathcalbf{G}^{(3)} \times^1_3 \cdots \times^1_3 \mathcalbf{G}^{(N)}$, where the set of $R_n$ for $n=0,\ldots,N$ and $R_0 = R_N = 1$ is referred to as the \textit{TT-rank}. The compression properties of TTD can be applied to significantly compress neural networks while maintaining comparable performance \cite{Novikov2015tnn}.

% from an exponential in the mode dimension $\prod_{n=1}^N I_n$ to a linear $\sum_{n=1}^N R_{n-1} I_n R_{n}$, which is highly efficient for high $N$ and low TT-rank. TTD has been used to significantly compress neural networks while maintaining comparable performance \cite{Novikov2015tnn}.
% \begin{equation}\label{eq:TTContractions}
% 	\mathcalbf{X} = \mathcalbf{G}^{(1)} \times^1_2 \mathcalbf{G}^{(2)} \times^1_3 \mathcalbf{G}^{(3)} \times^1_3 \cdots \times^1_3 \mathcalbf{G}^{(N)}	
% \end{equation}

\subsection{Graph Signal Processing}
\label{sec:GSP}

% \textbf{Signals on Graphs}. 
A graph, $\mathcal{G} = \{\mathcal{V}, \mathcal{E}\}$, is defined by a set of $N$ vertices (or nodes) $\textit{v}_n \subset \mathcal{V}$ for $n = 1, \ldots , N$, and a set of edges connecting the $n^{th}$ and $m^{th}$ vertex $\textit{e}_{n,m} = (\textit{v}_n, \textit{v}_m) \in \mathcal{E}$, for $n=1,\ldots,N$ and $m=1,\ldots,N$. A signal on a given graph is a defined by a vector $\textbf{f} \in \mathbb{R} ^ {N}$ such that $\textbf{f}: \mathcal{V} \rightarrow \mathbb{R}$, which associates a signal value to every node on the graph \cite{stankovic2019graph}. A graph can be fully described in terms of its weighted adjacency matrix, $\textbf{A} \in \mathbb{R} ^ {N \times N}$, such that $\textit{a}_{n, m} > 0$ if $\textit{e}_{n,m} \in \mathcal{E}$, and $\textit{a}_{n, m} = 0$ if $\textit{e}_{n,m} \notin \mathcal{E}$. The adjacency matrix can also be represented in its normalized form as $\tilde{\textbf{A}} = \textbf{D}^{\frac{1}{2}} \textbf{A} \textbf{D}^{\frac{1}{2}}$, where  $\textbf{D} \in \mathbb{R} ^ {N \times N}$ is the diagonal degree matrix such that $d_{n,n} = \sum_m \textit{a}_{n,m}$.

\textbf{Graph Shift Filters}
The weighted adjacency matrix can be used as a shift operator to filter signals on graphs. Such a graph filter represents a linear combination of vertex-shifted graph signals, which captures graph information at a local level \cite{stankovic2019graphII}. For example, the operation $\textbf{g} = (\textbf{I} + \textbf{A})\textbf{f}$ produces a filtered signal, $\textbf{g} \in \mathbb{R}^{N}$, such that $g_n = f_n + \sum_{m \in \Omega_n} a_{n, m} f_m$, where $\Omega_n$ denotes the $1$-hop neighbours that are directly connected to the $n$-th node. For $M$ graph signals stacked in a matrix form as $\textbf{F} \in \mathbb{R}^{N \times M}$, the resulting graph filter can be compactly written as $\textbf{G} = (\textbf{I} + \textbf{A})\textbf{F}$ \cite{stankovic2019graphII}. 

\subsection{Recurrent Graph Tensor Networks}
\label{sec:RGTN}

A Recurrent Graph Tensor Network (RGTN) \cite{xu2020recurrent} models sequential data through a time-based, multi-linear graph filter in tensor-network format.

\textbf{gRGTN}
A general RGTN (gRGTN) extracts a feature map $\textbf{Y} \in \mathbb{R}^{J_1 \times I_1}$ from sequential data, $\textbf{X} \in \mathbb{R}^{J_0 \times I_1}$, via the forward pass given by  $\textbf{Y} = \sigma(\mathcalbf{R} \times_{3,4}^{1,2} \textbf{W}^{(x)} \times_2^1 \textbf{X})$, where $\textbf{W}^{(x)} \in \mathbb{R}^{J_1 \times J_0}$ is the input weight matrix and $\mathcalbf{R} \in \mathbb{R}^{J_1 \times I_1 \times J_1 \times I_1}$ is the multi-linear time-graph filter. Specifically, $\mathcalbf{R}$ is defined as $\mathcalbf{R} = \tensorize(\textbf{I} + (\textbf{A} \otimes \textbf{W}^{(r)}))$, where $\textbf{I} \in \mathbb{R}^{J_1 I_1 \times J_1 I_1}$ is the identity matrix, $\textbf{A} \in \mathbb{R}^{I_1 \times I_1}$ is the time-vertex based graph adjacency matrix with $I_1$ time-steps represented as graph nodes, and the weight matrix $\textbf{W}^{(r)} \in \mathbb{R}^{J_1 \times J_1}$ models information propagation between successive time-steps over $J_1$ features.

\textbf{fRGTN}
The fast RGTN (fRGTN) is defined by approximating $\textbf{W}^{(r)} \approx \textbf{I}$ in gRGTN, which leads to a reduced forward pass, $\textbf{Y} = \sigma(\mathbf{R} \times_{2}^{2} \textbf{W}^{(x)} \times_2^1 \textbf{X})$, where $\mathbf{R} \in \mathbb{R}^{I_1 \times I_1}$ is a standard graph shift filter defined as $\mathbf{R} = (\textbf{I} + \textbf{A})$, as discussed in Section \ref{sec:GSP}. The lower parameter complexity over gRGTN is due to the identity approximation of $\textbf{W}^{(r)}$. 

\textbf{fRGTN-TT}
If the problem is inherently multi-modal, then the large dense layer matrices of the fRGTN can be tensorized and represented in the Tensor-Train format, as discussed in \cite{Novikov2015tnn}. This leads to the highly efficient fRGTN-TT model, which preserves the inherent multi-modality and has drastically lower parameter complexity.

\subsection{Reinforcement Learning}
In reinforcement learning (RL), at each time step \textit{t}, a RL agent observes the state $s_t$ of the environment, takes an action $a_t$ and receives a reward $r_t$ from the environment. The RL agent then learns a decision policy $\pi^*$ from the pairs $(a_t,r_t)$, that is optimal in maximizing the long-term reward $r=\sum_{t}r_t$.
\label{sec:Q-Learning}

\textbf{Q-Learning}
In Q-Learning, the goal is to learn a function $Q(s,a)$ that maps from the state $s$ and action $a$ to the expected accumulated reward. In other words, it estimates the reward that an agent will receive when it performs action $a$ under the environment state $s$. More precisely, the function Q that we aim to learn is given by: 
$Q_{\pi}(s,a)=E[R_t|s_t=s,a_t=a,\pi]$, with $\pi$ as the decision policy and $R_t$ the accumulated reward given by $R_t=\sum{_t}\gamma^{t'-t}r_{t'}$, where $\gamma<1$ is the discount factor used to give immediate rewards a higher value. The optimal Q-function that results in the best policy $\pi^*$ which maximizes the expected reward is given by $Q^*(s, a) = max _{\pi} Q_{\pi}(s, a)$. Within Q-Learning, this function is estimated iteratively using the Bellman equation, $Q^*(s, a)= r_t + \gamma max_{a'} Q^*(s', a')$, where $s'$ is the new state after taking action $a$ under state $s$.

\textbf{Deep Q-Learning}
A deep neural network in this context is trained by minimizing the difference between the two sides of the Bellman equation, using the loss function
$L(\theta_i)=E_{(s,a)\sim\rho(\cdot)}[(y-Q(s,a;\theta_i))^2]$, where $i$ denotes the training iteration and $\theta$ the weights of the deep Q-network. The training examples are in the form $(s,a,r,s')$, where $\rho(s,a)$ denotes their distribution, while $y$ is the prediction of $Q(s, a)$ given by the Bellman equation $y = r + \gamma max_{a'} Q^*(s', a'; \theta_{i-1})|(s, a)$. To alleviate the issue of non-stationary targets, experience replay is used, whereby past experiences are stored in a buffer, from which a batch is sampled at every time instant to train the network with back-propagation and stochastic gradient descent.

\textbf{Double Deep Q-Learning}
A notorious issue with the deep Q-Learning approach is its tendency towards instability and overestimation of action values. This is a result of the fact that both the current $Q$ value, $Q(s, a)$, and estimated $Q$ value, $y = r + \gamma max_{a'} Q^{*}(s', a')$, are computed by a single Q-network. To mitigate this issue, we employ Double Deep Q-Learning \cite{hasselt2015deep}, where a separate target network, $\Tilde{Q}$, is used to compute the estimated $Q$ value. The weights of $\Tilde{Q}$ are updated at the end of every training episode by hard copying the weights of $Q$.

\section{Multi-Graph Tensor Networks}
\label{sec:GTN}

The RGTN model introduced in \cite{xu2020recurrent} was developed to model time-series problems related to sequential data, and is only defined for a single graph domain. To make the concept suitable for applications beyond time-series and in a Big Data setting, we next generalize the results from \cite{xu2020recurrent} to develop a \textit{Multi-Graph Tensor Network}, which is capable of handling multi-modal data defined on multiple graph domains that is not limited to time-series modelling.

\subsection{General Multi-Linear Graph Filter}

The time-based multi-linear graph filter, $\mathcalbf{R}$, was developed in \cite{xu2020recurrent} to model time-series problems through a time-graph adjacency matrix that reflects the temporal flow of information, as discussed in Section \ref{sec:RGTN}. For this filter to be extended to other domains, the underlying graph topology needs to be modified. More generally, given a weighted graph adjacency matrix, $\textbf{A} \in \mathbb{R}^{I_1 \times I_1}$, we can construct a multi-linear graph filter in the tensor domain, $\mathcalbf{F} \in \mathbb{R}^{J_1 \times I_1 \times J_1 \times I_1}$, as
\begin{equation}
\mathcalbf{F} = \tensorize \left(\textbf{I} + \left(\textbf{A} \otimes \textbf{P} \right) \right)
\end{equation}
where the propagation matrix $\textbf{P} \in \mathbb{R}^{J_1 \times J_1}$ models the flow of information between neighbouring vertices (as opposed to successive time-steps in the RGTN case). This allows us to generalize the multi-linear graph filter $\mathcalbf{F}$ to any given graph domain of any data modality.

\subsection{General Multi-Graph Tensor Network}

Consider a multi-graph learning problem where the input is an order-($M+1$) tensor $\mathcalbf{X} \in \mathbb{R}^{J_0 \times I_1 \times I_2 \times \cdots \times I_M}$ with $J_0$ features indexed along $M$ physical modes $\{I_1, I_2, \ldots, I_M\}$, such that a graph $\mathcal{G}^{(m)}$ is associated with each of the $I_m$ modes, $m=1, \ldots, M$. For this problem, we define:
\begin{enumerate}
    \item $\mathcal{A} = \{ \textbf{A}^{(1)}, \textbf{A}^{(2)}, \ldots, \textbf{A}^{(M)}\}$, a set of adjacency matrices $\textbf{A}^{(m)} \in \mathbb{R}^{I_m \times I_m}$ constructed from the corresponding graphs $\mathcal{G}^{(m)}$.
    
    \item $\mathcal{W} = \{ \textbf{W}^{(1)}, \textbf{W}^{(2)}, \ldots, \textbf{W}^{(M)}\}$, a set of weight matrices $\textbf{W}^{(m)} \in \mathbb{R}^{J_m \times J_{m-1}}$ used for feature transforms, where $J_m$, for $m=1,\ldots,M$ controls the number of feature maps at every $m$.
    
    \item $\mathcal{P} = \{ \textbf{P}^{(1)}, \textbf{P}^{(2)}, \ldots, \textbf{P}^{(M)}\}$, a set of propagation matrices $\textbf{P}^{(m)} \in \mathbb{R}^{J_m \times J_m}$, modelling the propagation of information over the neighbours of the graph $\mathcal{G}^{(m)}$.
    
    %\item $\mathcal{S} = \{ \sigma^{(1)}, \sigma^{(2)}, \ldots, \sigma^{(M)}\}$, a set of point-wise non-linear activation functions.
\end{enumerate}

Using the above objects and an optional activation function, $\sigma(\cdot)$, we can now define the general \textit{Multi-Graph Tensor Network} (gMGTN) layer characterized by the following forward pass: 
\begin{equation}
\label{eq:gMGTN}
\mathcalbf{Y} = \sigma \left(\mathcalbf{F}^{(M)} \times_{3,4}^{1, M+1} \textbf{W}^{(M)} \times_2^1 \cdots \times_2^1 \mathcalbf{F}^{(2)} \times_{3,4}^{1, 3} \textbf{W}^{(2)} \times_2^1 \mathcalbf{F}^{(1)} \times_{3,4}^{1, 2} \textbf{W}^{(1)} \times_2^1 \mathcalbf{X} \right)
\end{equation}
where $\mathcalbf{F}^{(m)} = \tensorize(\textbf{I} + (\textbf{A}^{(m)} \otimes \textbf{P}^{(m)}))$. The so defined forward pass generates a feature map, $\mathcalbf{Y} \in \mathbb{R}^{J_M \times I_1 \times \cdots \times I_M}$, from the input tensor, $\mathcalbf{X}$, through a series of multi-linear graph filter and weight matrix contractions, which essentially iterates the graph filtering operation across all $M$ graph domains.

\subsection{Fast Multi-Graph Tensor Network}
\label{sec:sMGTN}

The gMGTN introduced in the previous section learns a propagation matrix, $\textbf{P}^{(m)}$, and a weight matrix, $\textbf{W}^{(m)}$, for each of the $M$ graphs. For simplicity, let $J_m = J$ for $m=1,\ldots,M$; this results in a parameter complexity of $\mathcal{O}(MJ^2)$, which is linear in the number of graphs, $M$, and quadratic in the size of feature maps, $J$. This hinders the performance of gMGTN, since computation can become intractable for high dimensional multi-graph problems. To that end, we develop the fast \textit{Multi-Graph Tensor Network} (fMGTN) as a low-complexity variant of the gMGTN.

Similar to \cite{xu2020recurrent}, we can reduce the parameter complexity of gMGTN by: (i) approximating $\textbf{P}^{(m)} \approx \textbf{I}$ for $m=1,\ldots,M$; and (ii) using one single weight matrix, $\textbf{W}^{(x)} \in \mathbb{R}^{J_1 \times J_0}$, for all of the graph domains, where $J_1$ controls the number of hidden units (feature maps). This allows us to define the fMGTN with the following reduced forward pass
\begin{equation}
\label{eq:sMGTN}
\mathcalbf{Y} = \sigma \left(\mathbf{F}^{(M)} \times_{2}^{M+1} \cdots \times_2^4 \mathbf{F}^{(2)} \times_2^3 \mathbf{F}^{(1)} \times_2^2 \textbf{W}^{(1)} \times_2^1 \mathcalbf{X} \right)
\end{equation}
where $\mathbf{F}^{(m)} = (\textbf{I} + \textbf{A}^{(m)})$ is a standard graph shift filter as discussed in Section \ref{sec:GSP}. In contrast to the gMGTN model, the proposed fMGTN does not have to learn $\textbf{P}^{(m)}$ or $\textbf{W}^{(m)}$, which reduces the parameter complexity of the forward pass to $\mathcal{O}(J^2)$ but at the cost of lower expressive power.

\begin{figure}[h!]
	\centering
	\includegraphics[width=0.75\linewidth]{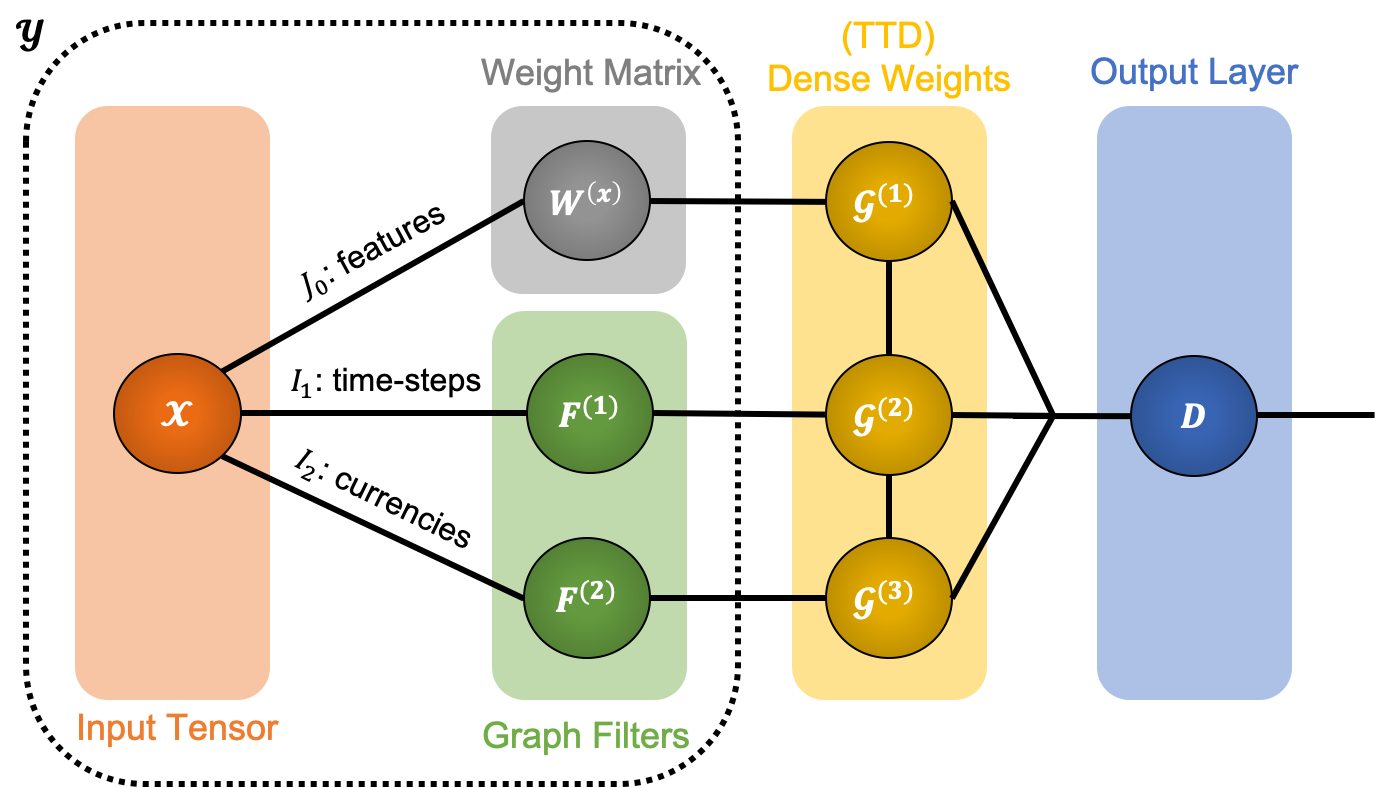}
	\caption{\textit{Tensor network representation of the fast Multi-Graph Tensor Network (fMGTN) used for our proposed experiment. The section encircled in dotted line denotes the multi-graph filtering operation for $M=2$ as per equation (\ref{eq:sMGTN}). The yellow region denotes a tensorized dense layer weight matrix, represented in the Tensor-Train format. The input data used for our experiment is an order-3 tensor with $J_0=4$ pricing features, $I_1=30$ past time-steps, and $I_2=9$ currencies, as discussed in Section \ref{sec:exp}. Note that we define a time-domain graph filter and a currency-domain graph filter for input data modes of dimensions $I_1$ and $I_2$, respectively.}}
	\label{fig:sMGTN}
\end{figure}

Finally, after extracting the feature map, $\mathcalbf{Y} \in \mathbb{R}^{J_1 \times I_1 \times \cdots \times I_M}$, it is customary to flatten the extracted features and pass them through dense layers to generate the final output. To further reduce the complexity, the weight matrices of the dense layers can be tensorized and represented in TT format, as discussed in \cite{Novikov2015tnn}. This further reduces the number of parameters, while maintaining compatibility with the inherent multi-modal nature of the problem. For clarity, an example of a fMGTN model which implements this series of contractions is shown in Figure \ref{fig:sMGTN}, using tensor network notation.

\section{Financial Environment}
\label{sec:FE}

\subsection{The Foreign Exchange Market and the Carry Graph}
\label{subsec:Carrygraph}

The FOREX market allows participants to trade pairs of currencies at a given \textit{spot rate}, which measures the value of a currency with respect to another currency at a given instant (e.g. EUR/USD spot rate of 1.2 implies that 1 Euro can be exchanged for 1.2 US Dollars). Alternatively, the participants can engage in forward contracts that allows them to exchange pairs of currencies on an agreed future date and at a specified \textit{forward rate}. If the forward rate of a given currency pair is higher than the current spot rate, then the numerator currency is expected to increase in value against the denominator currency and vice-versa. 

There are many factors that can affect the movements of spot rates, although the most important factor is arguably the \textit{carry} factor: a tendency for high interest rate currencies to generate higher returns than the low interest rate ones. According to the \textit{interest-rate-parity} theory \cite{aliber1973interest}, the expectation of currency pairs moving in different directions, depending on the interest rate differential, is reflected in the difference between the spot rate and the forward rate. 

\begin{figure}[h!]
	\centering
	\includegraphics[width=0.45\linewidth]{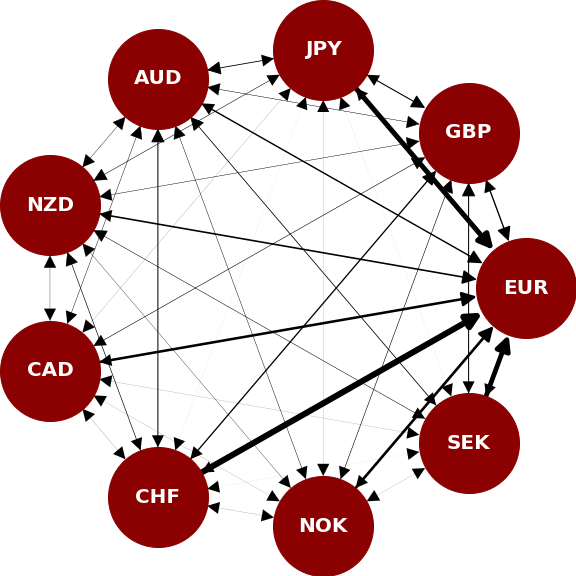}
	\caption{\textit{An example of FOREX carry graph. Thicker edges indicate stronger carry signal.}}
	\label{fig:carry_graph}
\end{figure}

Therefore, for a pair of currencies, $i$ and $j$, we can construct a pairwise carry signal by computing $c_{i,j}=1-\frac{r_f}{r_s}$, where $r_f$ and $r_s$ denote respectively the forward rate and the spot rate of the currency pair. Finally, we can construct a carry graph adjacency matrix $\textbf{A}$ such that its entries $a_{i,j}$ depend on the magnitude of the carry signal $c_{i,j}$. Figure \ref{fig:carry_graph} shows an example of the so constructed carry graph. 

\subsection{Characteristics of FOREX Data}

The FOREX data are characterized by a number of properties that make classical machine learning techniques inadequate for their modelling; these include:
\begin{itemize}
    \item FOREX data are multi-modal in nature, since they contain multiple pricing information indexed over time and across several related assets, which results in large dimensional tensors whose computation suffers from the Curse of Dimensionality.
    \item Financial data is known to have low signal-to-noise ratio due to the arbitrage forces in the market \cite{de2020machine}, which makes training particularly susceptible to overfitting.
    \item Various market factors can influence the pricing at different degrees depending on the time-horizon; this constitutes a multi-resolution problem that not many machine learning algorithms can handle.
\end{itemize}

The proposed MGTN is particularly suited to address the above challenges, as:
\begin{itemize}
    \item The multi-modal nature of FOREX data naturally leads to a tensor representation, which can be readily handled by the tensor network structure of our proposed model. 
    \item The model can leverage the powerful low-rank compression and regularization properties of tensor networks, which are inherently immune to the Curse of Dimensionality and provide a regularization framework via TD that does not degrade the underlying data structure.
    \item Long-term market factors such as \textit{carry} can be encapsulated in graph filters that naturally appeal to the pair-wise formulation of the FOREX data; this allows to process high frequency pricing data through an economically meaningful low-frequency graph topology.
\end{itemize}
% Specifically, the pair-wise formulation of FOREX data naturally leads to an economically meaningful graph structure, while its multi-modality is inherently suited for the regularization properties of the underlying tensor network that caters for the overfitting that naturally occurs in the extremely noisy FOREX market.  
\section{Experiments}
\label{sec:exp}

To demonstrate the applicability and superiority of the proposed framework for the task of algorithmic trading, we use the fMGTN model as a feature extraction part of the deep Q network of a trading agent, and evaluate its performance against three commonly used agents based on: (i) Gated Recurrent Unit (GRU) Neural Network \cite{Chung2014}, (ii) Tensor-Train Neural Network (TTNN) \cite{Novikov2015tnn}, and (iii) Graph Convolutional Network (GCN) \cite{kipf2016semi}. 

% \subsection{Experiment Settings}
% \paragraph{Data Description}
\textbf{Data Description}
Minute-wise spot-rate pricing data were used for the period between October 1\textsuperscript{st} 2019 and October 9\textsuperscript{th} 2019 for a total of 9 currencies. Training took place over the first 7 days, while out-of-sample performance evaluation was performed in the last 2 days. Our features include Open/High/Low/Close (OHLC) spot-rates of the 9 currencies presented in Figure \ref{fig:carry_graph}. 

\textbf{Data Pre-Processing}
The pricing data were pre-processed by computing the log-returns defined as $r_t = \ln(p_t)-\ln(p_{t-1})$, which measure the log-price difference between successive time-steps. The log-returns were then aggregated into multi-modal input samples, such that each sample, $\mathcalbf{X} \in \mathbb{R}^{J_0 \times I_1 \times I_2}$, contains log-returns indexed along $J_0=4$ features (OHLC), $I_1=30$ past time-steps, and $I_2=9$ currencies. Such multi-modal input data samples can be readily processed by the TTNN and the proposed fMGTN in their natural tensor form, as shown in Figure \ref{fig:sMGTN}. However, the input samples were matricized as $\mathbf{X} \in \mathbb{R}^{I_1 \times J_0I_2}$ and $\mathbf{X} \in \mathbb{R}^{I_2 \times J_0I_1}$ for compatibility with the GRU and the GCN agents, respectively.

\textbf{Graph Filter Formulation}
For the given FOREX input data, we formulate the problem as a multi-graph learning problem, where each sample contains $J_0$ features indexed along $I_1$ time-steps and $I_2$ currencies, whereby a time-graph and a currency-graph are associated with the time mode and the currency mode respectively. More precisely, we formulate the time-graph in the same way as in \cite{xu2020recurrent} for a total of $I_1=30$ time-steps, while the currency-graph was based on the carry-graph as discussed in Section \ref{subsec:Carrygraph}. Finally, the respective graph filter for our fMGTN model was computed as discussed in Section \ref{sec:sMGTN}. This results in graph filters $\textbf{F}^{(1)}$ and $\textbf{F}^{(2)}$, as illustrated in Figure \ref{fig:sMGTN}.

% \begin{table} [h!]
%   \caption{Agents' specifications}
%   \label{table:specs}
%   \centering
%   \begin{tabular}{llll}
  
%     \toprule
%      Agent & Layers & Units & Activation \\
%     \midrule
  
%   sMGTN & sMGTN & 16 & ReLU \\ & TT &  27 & ReLU    \\
%   \midrule
%   GRU & GRU & 16 & ReLU \\ 
%   & Dense & 27 & ReLU  \\
%   \midrule
%   TTNN & TT & 16 & ReLU   \\
%   & TT & 27 & ReLU \\
%   \midrule
%   GCN & GCN & 16 & ReLU \\
%   & Dense & 27 & ReLU \\
%     \bottomrule
%   \end{tabular}
% \end{table}
\textbf{Agent Specification}
For comparable results, the action value approximation network had the same specifications across all agents, with the sole difference being the feature extraction method. More specifically, each agent was based on a 3-layer architecture: (i) a feature extraction layer with 16 units and ReLU activation, (ii) a dense layer with 27 units and ReLU activation, and (iii) a linear output layer with 2 units corresponding to buy and sell action values.  Each agent uses a different feature extraction layer, which can be based on fMGTN, GRU, TTNN, or GCN. In addition, due to their inherent tensor representation, the dense layer following the fMGTN and TTNN layers was tensorized and represented in TT format. All agents were trained using ADAM with a learning rate of $2\cdot10^{-4}$ and a mini-batch size of 64 for 15 episodes. Finally, the reward of the agents consisted of minute-wise log-returns. Our models were implemented using Tensorflow 2.3
\footnote{github.com/gylx/GTNRL-Trading} \footnote{The GCN implementation was adapted from github.com/vermaMachineLearning/keras-deep-graph-learning}. 

\textbf{Performance Metrics}
Five different financial metrics were used to assess the performance of the agents: \textit{Total return} measures the total percentage returns generated by the agent in the episode; \textit{Sharpe ratio} measures the risk-adjusted return generated by the agent, which is computed as $\frac{\mu_r}{\sigma_r}$ where $\mu_r$ is the average log-return and $\sigma_r$ the standard deviation of log-returns; \textit{Sortino ratio} computes the risk-adjusted return as $\frac{\mu_r}{\sigma_r^{d}}$, where $\sigma_r^{d}$ measures the standard deviation of negative log-returns; \textit{Max Drawdown} measures the maximum percentage loss incurred by the agent during a consecutive period; \textit{Hit Rate} measures the percentage of profitable trades to total trades. 

\begin{figure}[h!]
	\centering
	\includegraphics[width=0.9\linewidth]{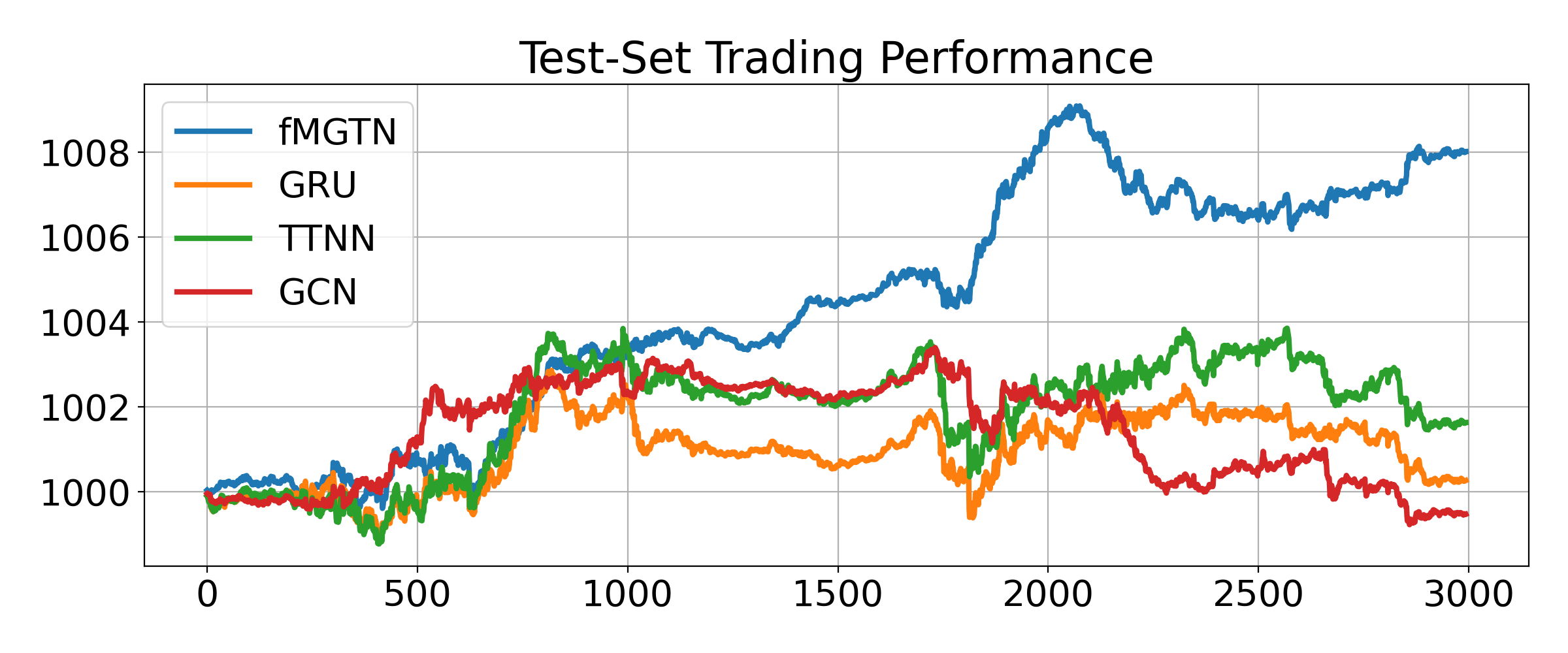}
	\caption{\textit{Out-of-sample trading performance of the considered agents, averaged over five European currencies. The vertical axis represents the investment growth of an initial value of 1000\$, while the horizontal axis represents time in minutes.}}
	\label{fig:performance}
\end{figure}

\textbf{Experimental Results}
Highly superior performance for the fMGTN based agent was obtained across a basket of European currencies, both in terms of generated profits and other common financial metrics. As shown in Figure \ref{fig:performance}, the fMGTN agent generated substantial profit (0.8\%) during the out-of-sample testing period. Table \ref{table:performance} summarizes the performance of the considered agents, with fMGTN  outperforming the other considered agents across a multitude of the most commonly used financial performance metrics. Finally, fMGTN achieved the best performance at a drastically lower parameter complexity, using up to 90\% less trainable parameters compared to the GCN agent, and up to 80\% less compared to the GRU agent, as shown in Table \ref{tab:mdl_complexity}. This confirms the potential of the proposed framework for integrating graphs, tensors, and neural networks, which proved superior to any of the three components acting individually. 

\newpage

\begin{table} [h!]
  \label{table:performance}
  \small
  \centering
  \begin{tabular}{llllll}
  
    \toprule
     Agent & Total Return (\%) & Sharpe Ratio & Sortino Ratio  & Max Drawdown (\%) & Hit Rate (\%) \\
    \midrule
  
   fMGTN & \textbf{0.8018} &  \textbf{0.0445} &  \textbf{0.0604} &\textbf{0.2893} & \textbf{52.8056} \\
   \midrule
   GRU & 0.0260 & 0.0012 & 0.0015 & 0.3477 & 50.4008 \\
   \midrule
   TTNN & 0.1628 & 0.0064 & 0.0083 & 0.3493 & 50.6346 \\
   \midrule
   GCN & -0.0538 & -0.0032 & -0.0040 & 0.4180 & 50.2338 \\
    \bottomrule
  \end{tabular}
  \vspace{5mm}
  \caption{\textit{Performance comparison for the considered agents}}
\end{table}

\begin{table}[h!]
    \centering
    \small
    \begin{tabular}{llll}
    \toprule
    fMGTN &  GRU & TTNN & GCN \\
    \midrule
    {531} & 3107 & 451 & 5891  \\
    \bottomrule
    \end{tabular}
    \vspace{5mm}
    \caption{\textit{Number of trainable parameters of the considered agents}}
    \label{tab:mdl_complexity}
\end{table}

\section{Conclusion}
\label{sec:conc}

We have introduced a novel framework that embarks upon the advantages of both graphs and tensors, to provide an efficient and meaningful modelling strategy in a deep learning setting. The so introduced \textit{Multi-Graph-Tensor-Network} (MGTN) has been shown to be capable of handling irregular data residing on multiple graph domains, while simultaneously leveraging on the compression properties of tensor networks to enhance the modelling power and drastically reduce parameter complexity. The effectiveness of the proposed model has been demonstrated on FOREX algorithmic trading, a challenging task owing to multiple sources of uncertainty and multi-modality. Experimental results have demonstrated the superiority of the proposed MGTN framework, which has generalized better in the highly irregular and noisy financial environment than the Recurrent Neural Network, Tensor-Train Neural Network, and Graph Convolutional Network based agents. 

Future research directions include work leveraging on the versatility of the proposed framework to investigate its potential in numerous applications that share the modelling setup considered here. For example, the graph filters within the MGTN allow for the modelling of irregular data defined on one or multiple graph domains, a typical setting in social networks, recommender systems, and traffic forecasting. In addition, the tensor network structure of the MGTN allows for the modelling of high-dimensional data at a low complexity, which appeals to problems including multi-sensor processing, video classification, and natural language processing. Future research on spectral MGTN models can also potentially improve the modelling power of MGTN through spectral graph filtering techniques.

\section*{Broader Impact}

The proposed multi-graph tensor network framework exploits the convergence of two successful but independently considered areas of graphs and tensor networks. This equips the MGTN framework with the ability to have considerable impact in applications where data sources come from irregular domains and are intrinsically high dimensional. This is a typical scenario in modern applications in finance, social networks, and physical sciences, to name but a few. 

An immediate impact of our work, as illustrated through the considered experiment, may be in the creation of successful FOREX algorithmic trading strategies that efficiently model the underlying variable coupling through the proposed multi-graph domain. Another potential outreach of our work is on the modelling of molecule structures through large graphs, that could in turn lead to more efficient drug discovery. 

The impact arising from model failure and data bias is identical to those of supervised learning models in general. Finally, we have not identified anyone that can be put at disadvantage from this work.

\begin{ack}
Y.L.X. is supported by an EPSRC Doctoral Scholarship. K.K. is supported by an EPSRC International Doctoral Scholarship.
\end{ack}

\appendix

\medskip
\small

\bibliographystyle{unsrtnat.bst}
\bibliography{references}

\end{document}